# Viewpoint Invariant Object Detector

## Graduation Thesis Extended Abstract

Osama Khalil        Andrew Habib**Introduction:**

Object Detection is the task of identifying the existence of an object class instance and locating it within an image. Difficulties in handling high intra-class variations constitute major obstacles to achieving high performance on standard benchmark datasets (scale, viewpoint, lighting conditions and orientation variations provide good examples). Suggested model aims at providing more robustness to detecting objects suffering severe distortion due to < 60° viewpoint changes. In addition, several model computational bottlenecks have been resolved leading to a significant increase in the model performance (speed and space) without compromising the resulting accuracy. Finally, we produced two illustrative applications showing the potential of the object detection technology being deployed in real life applications; namely content-based image search and content-based video search.

**Research Component**

**Approach**

Various features have been suggested in literature to be invariant under the aforementioned variations and thus providing good object class characteristics for detection. One of the most successful types of features is Scale-Invariant Feature Transform (SIFT) [1]. SIFT computes feature points in images invariant to change in lighting conditions, rotation and scaling. Many approaches have developed on top of SIFT increasing the limits of its invariance including Histogram of Oriented Gradients (HOG) in [2]. Object Detector models make use of such features in building discriminators/generators. Parts based models have attracted much attention lately in achieving state-of-art-results on VOC benchmark challenge [6]. Deformable part models treat objects as consisting of multiple connected parts with certain degrees of allowed deformations in space. Particularly, [3] is considered a major landmark in the development in such approach. [3] utilizes a variation of HOG to build a classifier using latent support vector machines. Yet, such features used are incapable of handling < 60° viewpoint variations as illustrated in [4] and studied by [5].

[4] suggests a method for selecting feature points that are invariant to < 60° viewpoint variations. The suggested method is analogous to the way SIFT selects feature points invariant to scaling. SIFT computes image features at different scales and selects

stable sharp points across scales. Affine-SIFT, as named by [4], computes SIFT image features in different orientations selecting the most stable feature points. While producing different scales of an image is a rather straightforward process, producing different viewpoints of the same scene captured in an image is less obvious. [4] shows that simple combined global and local affine transformations can produce the needed < 60° viewpoint variations. We follow the lines of [3] in utilizing feature points at all scales everywhere in the image (formally known as dense feature pyramid). We augment the feature pyramids computed with computed features in different orientations to achieve invariance under scale and viewpoint changes at the same time.

**Experimental Evaluation**

We tested the new suggested feature calculation technique using VOC 2010 challenge dataset containing over 20,000 images divided between training and testing. Training images are provided with ground truth annotations in the form of bounding boxes of the objects contained in each image. Testing set is not provided with annotations and evaluation is performed through an online server managed by PASCAL [6]. Due to limits on computational resources, we experimented with 3 out of the 20 suggested object classes; namely person, bicycle and car, used only 70% of the training set for each of the classes to train our model and didn't rescore our model's detections using contextual information (a step described in [3]). We set the deformable parts model parameters as suggested by [7] including SVM parameters and number of scales in the feature pyramid. Accuracy was calculated as the average precision over certain levels of recall. The following table shows our achieved results.

|  | [3] with full training set | Ours with 70% of training set |
|---|---|---|
| Person | 45.3 | 38.7 |
| Car | 38.6 | 33.5 |
| Bicycle | 45.3 | 32.1 |

We believe the reasons for not achieving matching numbers lie in the experimental restrictions imposed due to tight computational resources. Further experimental evaluation is needed to give assertions on the superiority of either approach.

**Technical Component**

Deploying more efficient garbage collection techniques and allowing for caching computed features to the hard disk, we managed to significantly decrease memory consumption and of the model in [7]. Achieving that allowed for creating the following illustrative applications:

<u>Content-based Image Search:</u> Given an image repository and an object class, our application returns the images that contain instances of the input object class, the instances being indicated by a bounding box.

<u>Content-based Video Search:</u> Given a video and an object class, our application returns the instants in the video that contain instances of the object class, the instances being bounded by a box. The application allows navigation to the video scene where the object starts to appear.